\documentclass[conference]{IEEEtran}
\IEEEoverridecommandlockouts
\usepackage{verbatim}

\usepackage{cite}
\usepackage{amsmath,amssymb,amsfonts}
\usepackage{algorithmic}
\usepackage{float}
\usepackage{graphicx}
\usepackage{textcomp}
\usepackage[dvipsnames]{xcolor}
\usepackage{enumitem}
\usepackage{dblfloatfix}
\usepackage{titlesec}
\usepackage[square,numbers]{natbib}
\usepackage{url}
\definecolor{mygreen}{RGB}{0, 170, 0}
\definecolor{mygreen_time_graph}{RGB}{0, 128, 0}

\usepackage{tikz}
\newcommand*\circled[1]{\tikz[baseline=(char.base)]{
            \node[shape=circle,color=black, draw,inner sep=0.8pt, minimum size=2pt] (char) {#1};}}
\newcommand{\rpoint}[1]{\circled{{\fontfamily{pcr}\selectfont\footnotesize{#1}}}}

\titlespacing\section{0pt}{0.3\baselineskip}{0.2\baselineskip}
\titlespacing\subsection{0pt}{0.3\baselineskip}{0.1\baselineskip}
\titlespacing\subsubsection{0pt}{0.2\baselineskip}{0.1\baselineskip}

\def\BibTeX{{\rm B\kern-.05em{\sc i\kern-.025em b}\kern-.08em
    T\kern-.1667em\lower.7ex\hbox{E}\kern-.125emX}}

\usepackage{fancyhdr,lipsum}
\fancypagestyle{firstpage}{
  \fancyhf{}
  \fancyhead[C]{To appear at the 2021 International Joint Conference on Neural Networks (IJCNN), Virtual Event, July 2021.}
  \fancyfoot[C]{\thepage}
}

\begin{document}
\title{CarSNN: An Efficient Spiking Neural Network for Event-Based Autonomous Cars on the Loihi Neuromorphic Research Processor\\
\vspace*{-10pt}}


\author{\IEEEauthorblockN{Alberto Viale$^{1,2,*}$\thanks{*These authors contributed equally to this work.}, Alberto Marchisio$^{1,*}$, Maurizio Martina$^2$, Guido Masera$^2$, Muhammad Shafique$^3$}
\IEEEauthorblockA{\textit{$^1$Technische Universit{\"a}t Wien, Vienna, Austria}\ \ \ \textit{$^2$Politecnico di Torino, Turin, Italy}\ \ \ \textit{$^3$New York University, Abu Dhabi, UAE}} 
\IEEEauthorblockA{\small{\textit{Email: \{alberto.viale, alberto.marchisio\}@tuwien.ac.at, \{maurizio.martina, guido.masera\}@polito.it, muhammad.shafique@nyu.edu}}\\
}\vspace*{-30pt}}

\maketitle
\thispagestyle{firstpage}





\begin{abstract}


Autonomous Driving (AD) related features provide new forms of mobility that are also beneficial for other kind of intelligent and autonomous systems like robots, smart transportation, and smart industries. For these applications, the decisions need to be made fast and in real-time. Moreover, in the quest for electric mobility, this task must follow low power policy, without affecting much the autonomy of the mean of transport or the robot. These two challenges can be tackled using the emerging Spiking Neural Networks (SNNs). When deployed on a specialized neuromorphic hardware, SNNs can achieve high performance with low latency and low power consumption. In this paper, we use an SNN connected to an event-based camera for facing one of the key problems for AD, i.e., the classification between cars and other objects. To consume less power than traditional frame-based cameras, we use a Dynamic Vision Sensor (DVS)~\cite{DVS_camera}. The experiments are made following an offline supervised learning rule, followed by mapping the learnt SNN model on the Intel Loihi Neuromorphic Research Chip~\cite{Loihi_chip}. Our best experiment achieves an accuracy on offline implementation of 86\%, that drops to 83\% when it is ported onto the Loihi Chip. The Neuromorphic Hardware implementation has maximum 0.72 ms of latency for every sample, and consumes only 310 mW. To the best of our knowledge, this work is the first implementation of an event-based car classifier on a Neuromorphic Chip. 

\end{abstract}

\begin{IEEEkeywords}
Autonomous Driving, AD, Spiking Neural Networks, SNN, Spatio-Temporal Backpropagation, STBP, Intel Loihi, Neuromorphic Computing, Dynamic Vision Sensor, DVS, cars vs. background classification.
\end{IEEEkeywords}

\section{Introduction}


The interest in Autonomous Driving (AD) has significantly grown in recent years. Therefore, new algorithms and design solutions have to address the challenges offered by this rapidly expanding sector. This paper focuses on practical AD systems by proposing a Spiking Neural Network that is directly implementable on a Neuromorphic Chip using a DVS, which can be easily introduced inside the car control system.

\subsection{Target Research Problem and Research Challenges}

If we study the various driving operations for a vehicle, we can understand how large/complex the AD problem space is, and how difficult it is to consider the problem in its entirety. For the purposes of this research, we can separate the AD tasks in the following three principal categories:
\begin{itemize}[leftmargin=*]
    \item the classification of the environment objects such as pedestrian and cars~\cite{N-cars_dataset};
    \item the prediction of the position of these objects~\cite{GEN1};
    \item the prediction of the controls of the car such as the steering angle and the status of the brake pedal and accelerator~\cite{DDD20}.
\end{itemize}

These can be viewed as regression and generalization problems, because with some different inputs coming from the sensors, the AD system has to predict a reaction that represents the solution for the task. The main methods to make these decisions in fast and accurate ways are represented by Deep Learning algorithms, that are typically divided into the following types:
\begin{itemize}[leftmargin=*]
    \item \textbf{Deep Neural Networks (DNNs)}, that are the oldest and can be implemented on traditional hardware processors and specialized architectures~\cite{Sze2017EfficientDNN}\cite{Capra2020SurveyDNN}. They are based on the transmission of digital values, but exhibit high power consumption.
    \item \textbf{Spiking Neural Networks (SNNs)}, that closely follow the behavior of neurons and are based on the transmission of spikes. They can be implemented on conventional hardware, but to achieve very low power consumption they are more amenable to the Neuromorphic Chips~\cite{Schuman2017SurveyNeuromorphic}. This aspect can be also optimized by the implementation on energy-efficient frameworks like SparkXD~\cite{SparkXD}, FSpiNN~\cite{FSpiNN}, Q-SpiNN~\cite{Putra2021QSpiNN} or SpikeDyn~\cite{spikedyn}.
\end{itemize}

Since every task represents a real-time problem, we want that the entire decision-making system has a good reactivity with a very low latency, in order to minimize the chance to have catastrophic car accidents due to late decisions. Another challenge is related to the robustness of the system that must operate in all conditions, in particular different types of illumination and weather conditions. Moreover, the system design should be optimized for low power consumption, which is an important design criteria for automotive, especially in the battery-driven electric mobility.

In our research, we focus on the \textit{``cars vs. background''} classification problem. 
To overcome the above-discussed limitations, we identify three main research objectives:
\begin{enumerate}[leftmargin=*]
    \item the system should use the major robust vision engine, i.e., an event-based camera;
    \item the network should be a low-complexity event-based SNN for energy-constrained systems;
    \item the developed SNN should fulfill the system constraints to be implemented onto a neuromorphic hardware chip.
\end{enumerate}
Following these research targets, we design, optimize, and implement the SNN on the Intel Loihi Neuromorphic Research Chip~\cite{Loihi_chip}, and evaluate it on the N-CARS dataset~\cite{N-cars_dataset}. It is based on Asynchronous Time-based Image Sensor (ATIS)~\cite{ATIS_camera}, which is an event-based camera.  

\subsection{Motivational Case Study}


Since, to the best of our knowledge, there were no prior existing works on AD applications implemented onto the Loihi Neuromorphic Chip, to highlight the research problems, we provide a motivational case study by analyzing another application. A well established benchmark in the event-based neuromorphic community is constituted by the \textit{IBM DVS128 gesture dataset}~\cite{GESTURE_DATA}, which is a database for gesture classification. For example, the work in~\cite{Motivational_case} uses an SNN trained offline on this dataset to recognize live operator gestures taken by a DVS camera connected to the \textit{Intel Kapoho Bay} Neuromorphic Chip. It achieves about $91\%$ accuracy, having also the possibility to learn online new gestures with few shots, using the On-Chip Learning engine. In the last case it achieves about $80\%$ accuracy results with only 20 shots.

\begin{figure}[h]
    \centering
    \includegraphics[width=\linewidth]{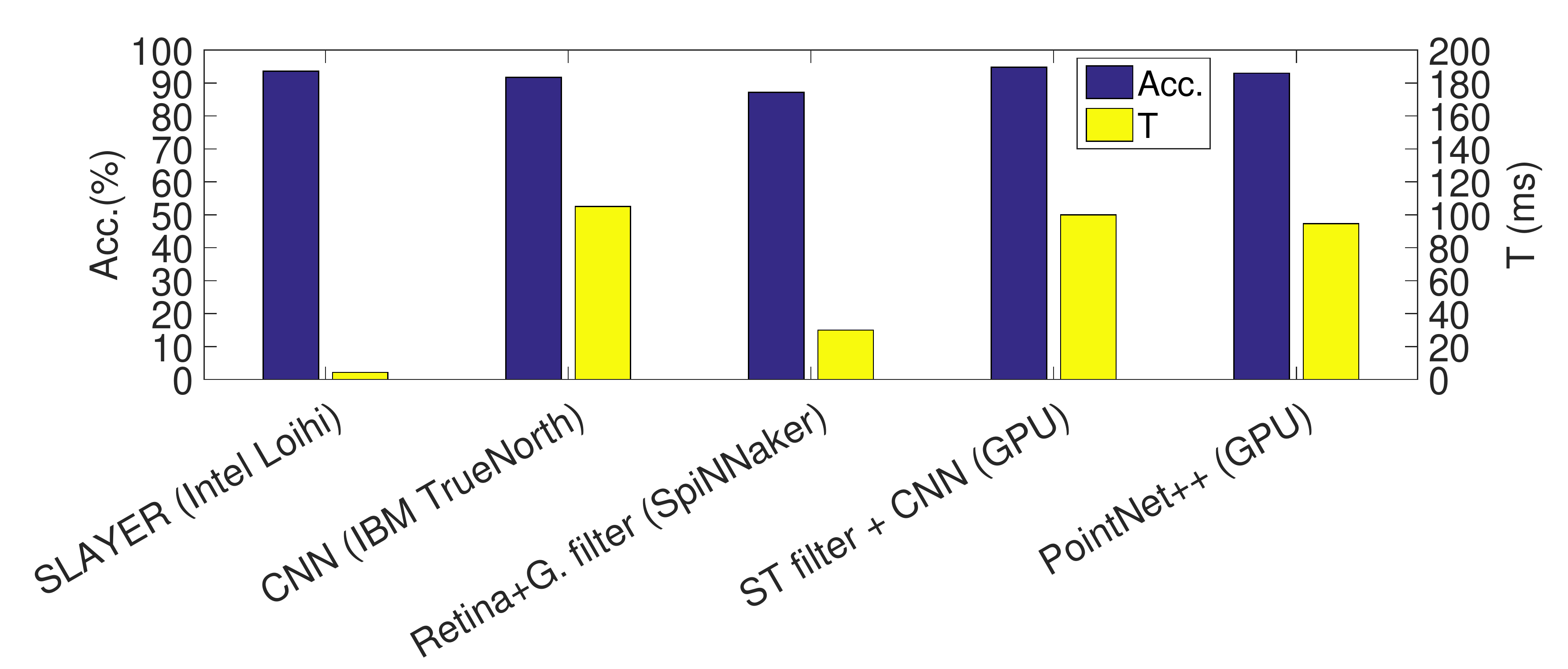}
    \caption{Accuracy (Acc.) and latency (T) comparison between different implementations for DVSGesture recognition problem~\cite{Slayer_shrestha,TrueNorth,retina,spatiotemporal,PointNet}. The higher accuracy to latency ratio is achieved by the SLAYER implemented on the Intel Loihi. It also has the lowest power consumption (0.54 mJ), compared to over 19 mJ consumed by the IBM TrueNorth implementation.}
    \label{fig:motivational_comparison}
\end{figure}

Figure~\ref{fig:motivational_comparison} compares the implementations of classifiers on different hardware platforms, in terms of accuracy and latency. The SLAYER implementation on Loihi~\cite{Slayer_shrestha} exhibits the shortest latency (only $4.35\ ms$), with accuracy comparable with an optimized GPU implementation~\cite{spatiotemporal}. Therefore, towards real-time use cases, these results motivate us to conduct this research on AD applications on the Loihi chip.


\subsection{Our Novel Contributions}
In this paper we present \textbf{CarSNN}, a novel spiking Convolutional Neural Network (CNN) based classifier method to tackle the classification problem between cars and background images collected by an event-based camera.


Using the attention window strategy, the focus is concentrated only on a part of the original input. To find the best window size and position, we analyze the statistics of input spikes, and focus on the part with more information (\textbf{Section~\ref{section:analysis}}).

To maintain the temporal correlation between different events we also adopt the accumulation in time of these information (\textbf{Section~\ref{section:Methodology}}). We use three hierarchical stages for this strategy implementation:

\begin{enumerate}[leftmargin=*]
    \item During a time window, called \textit{sample time}, the spikes from the event-based camera are collected following the rule for which each channel can have a maximum of one spike per pixel coordinate.
    \item The resulting image is given to the input of the network and remain stable for many time steps. We take the class with highest output spikes as the prediction of this single image. 
    \item To increase the accuracy we can derive more than only one image from the sample stream. To do that, we define a second time window, called \textit{sample length}, that is multiple of \textit{sample time}. Therefore, we have \textit{sample length / sample time} different input images from a single sample stream. Based on the classification made for every single image, the most predicted class represents the prediction for the entire sample stream.
\end{enumerate}

Moreover, with the above-discussed steps, we obtain a compression of the information, which is extremely important for low power applications.

All the developed networks followed the constraints of an existing Neuromorphic Chip, the Intel Loihi, to face the cars classification problem with event-based camera streams (\textbf{Section~\ref{section:result}}).

To give an overview of every technical part used to perform the task, in \textbf{Section~\ref{section:bg_related_work}} we:

\begin{itemize}[leftmargin=*]
    \item focus on what an \textit{event-based vision sensor} is and what are the advantages over frame-based cameras (\textbf{Section~\ref{subsection:Event-based camera}});
    \item present the \textit{Neuromorphic-cars} (\textit{N-CARS}) dataset, that can be used to train an SNN for the current classification problem (\textbf{Section~\ref{subsection:N-cars_data}});
    \item expose what a \textit{Spiking Neural Network} is, its advantages and how it can be modeled and trained (\textbf{Section~\ref{subsection:SNN}});
    \item summarize the main features and advantages of the \textit{Spatio-Temporal BackPropagation} Supervised Learning rule used in this paper (\textbf{Section~\ref{subsection:STBP}});
    \item explain the behavior of the \textit{Intel Loihi Neuromorphic Chip} used to find the major network constraints (\textbf{Section~\ref{Loihi_section}}).
\end{itemize}

For reproducible research, the source code for training and deploying our \textit{CarSNN} models has been released at \url{https://github.com/albertopolito/CarSNN}.

\section{Background and Related Work} \label{section:bg_related_work}

\subsection{Event-Based Cameras} \label{subsection:Event-based camera}
In recent years, the event-based cameras~\cite{DVS_camera}, which are bio-inspired sensors for the acquisition of visual information, were proposed and designed to overcome the performance of the classic frame-based cameras.
They recognize the same matrix of pixels, but they collect the information in a different way.
\begin{itemize}[leftmargin=*]
    \item A \textbf{frame-based camera} records the video as set of images and every image is collected with a constant delay from the neighbor in time, without any compression.
    \item An \textbf{event-based camera} records the video as a set of events. If and only if a pixel changes its brightness, the camera triggers an event with these information:
    \begin{itemize}[leftmargin=*]
        \item $x,y$: the coordinates of the pixel;
        \item $t$: the timestamp of when the event occurred;
        \item $p$: the polarity of the variation of the brightness, which is $ON$ or $1$ if the pixel is brighter, and $OFF$ or $0$ if the brightness is reduced.
    \end{itemize}
\end{itemize}

For their structure, event-based cameras are extremely useful when coupled with the SNNs, because the spikes generated by the sensor can directly feed the SNNs' inputs.\\
Event-based cameras have others advantages:
\begin{itemize}[leftmargin=*]
    \item \textit{High resolution in time}: it can record two different events delayed by few microseconds. Therefore, it does not suffer from oversampling, undersampling, and motion blur.
    \item \textit{Adaptive data rate and less memory usage}: there is no need to store the redundant information, but only the changes, to obtain an efficient storage of the information.
    \item \textit{High dynamic range} (up to $120 dB$): it can record scenarios with a great change of brightness without losing any information.
\end{itemize}

The major drawback of this camera is its lower resolution in space than a frame-based camera.

\subsection{N-CARS Dataset} \label{subsection:N-cars_data}

The main event-based datasets derive from a simulation of an event-based camera on frame-based recording images~\cite{event_base_frame1}\cite{event_frame2}. Hence, these benchmarks lose the great time bandwidth of event-based cameras~\cite{DVS_camera}. Introduced to overcome the limited numbers of event-based data recorded by an event-based camera from the real word, the N-CARS dataset~\cite{N-cars_dataset} is a recording of $80\ minutes$ with an ATIS camera~\cite{ATIS_camera}. This sensor has a resolution in space of 304$\times$240 and it is mounted behind the windshield of a car. For recognition purposes, the outcoming events are transformed into grey-scale images. These are processed with a state-of-the-art object detector~\cite{yolo9000}\cite{RCNN}, to automatically extract bounding boxes around the two classes:
\begin{itemize}[leftmargin=*]
    \item \textbf{cars}: $12336$ samples;
    \item \textbf{background}: $11693$ samples.
\end{itemize}

The maximum bounding boxes size is of $120 \times 100$ pixels.

\begin{figure}[h]
    \centering
    \includegraphics[width=\linewidth]{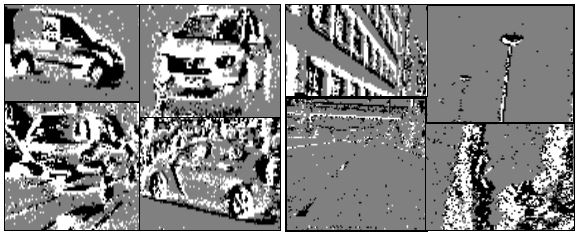}
    \caption{Example of grey-scale accumulated images of the N-CARS dataset~\cite{N-cars_dataset}. The four pictures on the left represent car samples, while the images on the right represent background samples.}
    \label{fig:N-cars}
\end{figure}

The dataset is split into 7940 car and 7482 background training samples;  4396 car and 4211 background testing samples. Each example lasts 100 milliseconds. 
The dataset files are grouped by class and are made as $1\ channel$ stream with two possible event values $-1$ and $1$. 
An example of the accumulated grey-scale images is shown in Figure~\ref{fig:N-cars}.

\subsection{Spiking Neural Networks (SNNs)}
\label{subsection:SNN}

As previously discussed, the \textbf{Spiking Neural Networks} (\textbf{SNNs}) introduce a revolution in the Artificial Intelligence and Machine Learning field~\cite{SNN_general}, since they are known as the third generation of neural networks. An SNN is the result of the research of what the \textit{neurons} effectively do into the brain. It is based on the most biological probable behavior, where the information is encoded into \textit{spikes} and spread through the neurons by \textit{axons} and \textit{synapses}. 
The behavior of the brain can be simulated with many kinds of models which can be more or less complex. For example, a complex model exhibits great performance, but on the other hand it is difficult to implement, due to high latency and high power consumption.  


The SNNs offered many advantages w.r.t. the older (non-spiking) DNNs~\cite{SNN_vs_ANN}:
\begin{itemize}[leftmargin=*]
    \item \textit{low power consumption}, due to the adaptation of the consumption with the intensity of the inputs;
    \item \textit{straightforward interface of event based sensors}, for example with DVS cameras as inputs to the system;
    \item \textit{low computation latency}, due to the asynchronous computation of the spikes and the speed of their spread. 
\end{itemize}

The most simple spiking neuron model is the \textbf{Integrate and Fire} (\textbf{IF}) model~\cite{IF_model}. It is based on the idea that every neuron can be represented by a resistance-conductance (RC) equivalent circuit.

A little more complicated, but also more biological plausible model is represented by a modified version of the IF model, which is called \textbf{Leaky-Integrate and Fire} (\textbf{LIF}) model~\cite{IF_model}.

\begin{figure}[h]
    \centering
    \includegraphics[scale=0.5]{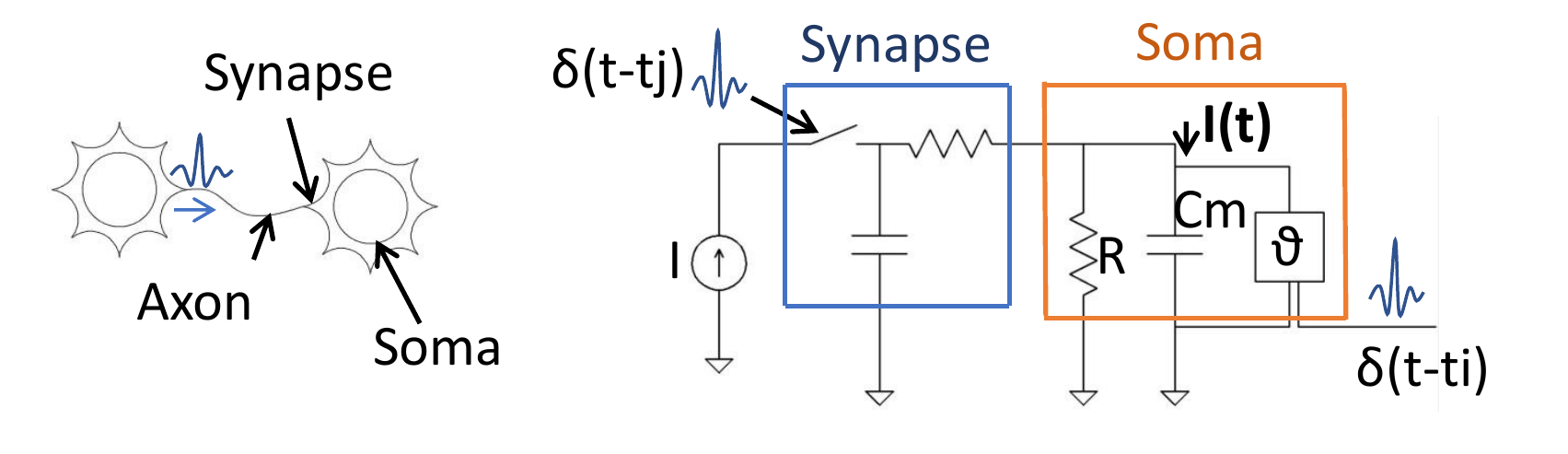}
    \vspace*{-6pt}
    \caption{Circuital representation of the \textbf{LIF} model for Post-synaptic Neuron that receives a spike sent by the Pre-synaptic Neuron~\cite{IF_model}.}
    \label{fig:IF_model}
\end{figure}

Figure~\ref{fig:IF_model} illustrates the model behavior of the interaction between two Neurons, called Pre-synaptic and Post-synaptic neurons. The \textit{pre-synaptic meuron} connects its \textit{axon} to the \textit{dentrites} (\textit{soma} in the figure) of the \textit{post-synaptic neuron} through a \textit{synapse}. The \textit{synapse} is represented by a low-pass filter, while the \textit{dentrite} or \textit{soma} is represented as a reservoir of charge, i.e., a capacitance. When the reservoir state, called \textbf{Post-synaptic Membrane Potential} (PSP) overcomes the threshold ($\theta$) the \textit{neuron} fires a spike through its \textit{axons}, and the reservoir state (PSP) resets to a value that is always less than the threshold $\theta$ (it can be 0 or a positive value). After that time, if there are others input spikes, the potential can increase again. This behavior can be modeled by the following differential Equation~\ref{eq:diffeq1}:
\begin{equation}
    I(t)=\frac{u(t)}{R}+C \frac{d v}{d t}
    \label{eq:diffeq1}
\end{equation}

The \textbf{Membrane Time Constant} \textbf{$\tau_m$} is derived with Equation~\ref{eq:tau_m}:
\begin{equation}
    \tau_{m} \frac{d u}{d t}=-u(t)+R I(t),
    \label{eq:tau_m}
\end{equation}
where $u(t)$ is the neuronal membrane potential at time $t$ and $I(t)$ denotes the pre-synaptic input, which is determined by the weighted pre-neuronal activities.

Here, in case of no incoming spikes from the synapses of the neuron, the PSP decreases over time by a fraction called $voltage\ decay$. This model also introduces the concept of $refractory\ period$, that is a short time, after the input of a spike from the synapses, in which the neuron is unable to consider others spikes at the input, which are then discarded. 
Every synapse of the neuron has a \textit{weight} that multiplies the incoming spikes before it can affect the PSP~\cite{Learning_methods}. This is the key feature of the generalization and regression mechanisms of the SNNs. 

To realize a given task accurately, the SNNs' weights need to be properly adjusted through a \textbf{Learning Process}. During this step, the SNN is fed by the train inputs of the dataset. With some method (\textbf{Learning Method}), that can be different according to the desired application, the weights are adjusted, in order to hit the target.\\
The SNN learning methods today are grouped in three main classes~\cite{Learning_methods}:
\begin{enumerate}[leftmargin=*]
    
    \item \textit{Direct supervised}: the SNN is stimulated with different patterns of spike trains and the synaptic weights are adjusted to achieve the desired output spike trains. The most common algorithms are based on back-propagation mechanism~\cite{STBP, STCA}.
    \item \textit{Indirect supervised}: a DNN is trained and then converted into an equivalent SNN~\cite{Massa2020EfficientSNN, Spike_train_learning_method}.
    \item \textit{Unsupervised}: the SNN is stimulated with a pattern of spike trains, but no human-produced labels are given. The SNN by itself searches the correlation properties between every sample. Unsupervised SNN learning can involve tasks such as cluster analysis~\cite{STDP} and anomaly detection~\cite{anomaly_detection}.
\end{enumerate}

Focusing on the \textit{direct supervised} method and in particular on the SNN back-propagation, there are two main problems:
\begin{itemize}[leftmargin=*]
        \item  $Spike\ as\ activation\ function$: since the SNN is based on spikes, i.e., impulses, the derivative of an impulse does not exist. The possible solution can be its approximation (e.g., a \textit{surrogate gradient})~\cite{surrogate_model}, but its implementation detaches from the biological model. However, with this solution many different learning rules can be applied and the SNN can also achieves high performance~\cite{STBP, STCA}.
        \item $Weight\ transport\ problem$~\cite{Learning_methods}: the SNN needs to have two paths, one for forward and one for backward. In this situation, the weights for these paths are correlated, being one the transposition of the other. This coherence is hard to maintain. One solution is to have random weights on the backward path, but this can be used only for simple problems~\cite{SpikeProp}. 
\end{itemize}

\subsection{Spatio-Temporal Back-Propagation Learning Method} \label{subsection:STBP}

The most common SNN learning method based on Back-propagation is the Spatio-Temporal Back-Propagation (\textbf{STBP})~\cite{STBP}. 
The most used supervised learning rules control the interaction between neurons (\textbf{Spatial Domain} or \textbf{SD}) in order to find new weights. On the other hand, the unsupervised methods monitor the trend over time of the neurons' PSP (\textbf{Time Domain} or \textbf{TD}), to do their tasks. The STBP uses both information to train the SNNs, thus using the back-propagation on both dimensions. 
The STBP algorithm starts from the LIF neuron model (Equation~\ref{eq:tau_m}) and resolves this TD differential problem to obtain Equation~\ref{eq:STBP_resolve_diff}:
\begin{equation}
   u(t)=u(t_{i-1})e^{\frac{t_{i-1}-t}{\tau}}+RI(t)
    \label{eq:STBP_resolve_diff} 
\end{equation}
 
In this way, both the TD and the SD components are present in the STBP method. $I(t)$ represents the \textit{spatial accumulation} and $u(t_{i-1})$ represents the \textit{leaky temporal memory}. Then, since the back-propagation algorithm takes many advantages from the iterative representation of the gradient descent, the authors of~\cite{STBP} developed iterative LIF-based SNNs, in which the iterations occur in both the SD and TD as follows (Equation~\ref{eq:system_STBP_model}):
\begin{equation}
\begin{aligned}
x_{i}^{t+1, n} &=\sum_{j=1}^{l(n-1)} w_{i j}^{n} o_{j}^{t+1, n-1} \\
u_{i}^{t+1, n} &=u_{i}^{t, n} f\left(o_{i}^{t, n}\right)+x_{i}^{t+1, n}+b_{i}^{n} \\
o_{i}^{t+1, n} &=g\left(u_{i}^{t+1, n}\right),
\end{aligned}
\label{eq:system_STBP_model}
\end{equation}
where:
\begin{equation}
f\left(o_{i}^{t, n}\right) \approx\left\{\begin{array}{ll}
\tau, & o_{i}^{t, n}=0 \\
0, & o_{i}^{t, n}=1
\end{array}\right.\ for\ little\ \tau
\label{eq:system_STBP_model_f}
\end{equation}

\begin{equation}
g(x)=\left\{\begin{array}{ll}
1, & x \geq V_{t h} \\
0, & x<V_{t h}
\end{array}\right.
\label{eq:system_STBP_model_g}    
\end{equation}

The notations of the above formulas (Equations~\ref{eq:system_STBP_model},~\ref{eq:system_STBP_model_f} and~\ref{eq:system_STBP_model_g}) are the following:
\begin{itemize}[leftmargin=*]
    \item the index $t$ represents the current time step;
    \item $n$ and $l(n)$ denote the $n^{th}$ layer and its number of neurons, respectively;
    \item $w_{ij}$ is the synaptic weight between the $j^{th}$ pre-synaptic neuron and the $i^{th}$ post-synaptic neuron;
    \item $o_j$ is the neuronal output of the $j^{th}$ neuron;
    \item $x_i$ is the pre-synaptic input of the $i^{th}$ neuron;
    \item $b_i$ is the bias of the $i^{th}$ neuron.
\end{itemize}

The learning rule defines a \textbf{Loss Function} (Equation~\ref{eq:loss_STBP}) and a \textit{Gradient Descendent Optimization Method} that consists of minimizing the loss function under a given time window $T$, using its derivative.

\begin{equation}
L=\frac{1}{2 S} \sum_{s=1}^{S}\left\|y_{s}-\frac{1}{T} \sum_{t=1}^{T} o_{s}^{t, N}\right\|_{2}^{2}
\label{eq:loss_STBP}    
\end{equation}

where $y_{s}$ is the label and $o_{s}$ is the output of the network for the $s^{th}$ sample.

Then, four different cases to perform the calculation of Equation~\ref{eq:system_STBP_derivative_general} are distinguished:
\begin{equation}
\delta_{i}^{t, n}=\frac{\partial L}{\partial o_{i}^{t, n}}    \label{eq:system_STBP_derivative_general}
\end{equation}

\begin{enumerate}[leftmargin=*]
    \item $t=T\ and\ n=N\ (output\ layer)$: 
    \begin{equation}
    \frac{\partial L}{\partial u_{i}^{T, N}}=\delta_{i}^{T, N} \frac{\partial o_{i}^{T, N}}{\partial u_{i}^{T, N}}    \label{eq:system_STBP_derivative_general_0}
    \end{equation}
    
    \item $t=T\ and\ n<N\ (inner\ layer)$:
    \begin{equation}
      \frac{\partial L}{\partial u_{i}^{T, n}}=\delta_{i}^{T, n} \frac{\partial g}{\partial u_{i}^{T, n}}  \label{eq:system_STBP_derivative_general_1}
    \end{equation}
    
    \item $t<T\ and\ N=n\ (output\ layer)$:
    \begin{equation}
      \frac{\partial L}{\partial u_{i}^{t, N}}=\delta_{i}^{t+1, N} \frac{\partial g}{\partial u_{i}^{t+1, N}} f\left(o_{i}^{t, n}\right)  \label{eq:system_STBP_derivative_general_2}
    \end{equation}
    
    \item $t<T\ and\ N<n\ (inner\ layer)$:
    \begin{equation}
    \begin{aligned}
\frac{\partial L}{\partial u_{i}^{t, n}} &=\delta_{i}^{t, n} \frac{\partial g}{\partial u_{i}^{t, n}}+\delta_{i}^{t+1, n} \frac{\partial g}{\partial u_{i}^{t+1, n}} f\left(o_{i}^{t, n}\right)
\end{aligned}    \label{eq:system_STBP_derivative_general_3}
    \end{equation}
    
\end{enumerate}

Afterwards, these differential equations (Equations~\ref{eq:STBP_diff_eq_end_1} and~\ref{eq:STBP_diff_eq_end_2}) can be defined:
\begin{equation}
\frac{\partial L}{\partial \boldsymbol{b}^{n}}=\sum_{t=1}^{T} \frac{\partial L}{\partial \boldsymbol{u}^{t, \boldsymbol{n}}} \frac{\partial \boldsymbol{u}^{t, \boldsymbol{n}}}{\partial L \boldsymbol{b}^{n}}=\sum_{t=1}^{T} \frac{\partial L}{\partial \boldsymbol{u}^{t, n}}
    \label{eq:STBP_diff_eq_end_1}
\end{equation}

\begin{equation}
    \frac{\partial L}{\partial \boldsymbol{W}^{n}}=\sum_{t=1}^{T} \frac{\partial L}{\partial \boldsymbol{u}^{t, n}} \frac{\partial \boldsymbol{u}^{t, n}}{\partial x^{t, n}} \frac{\partial x^{t, n}}{\partial W^{n}}=\sum_{t=1}^{T} \frac{\partial L}{\partial \boldsymbol{u}^{t, n}} \boldsymbol{o}^{t, n-1}
    \label{eq:STBP_diff_eq_end_2}
\end{equation}

such that they can be used to perform the \textbf{Gradient Descendent Optimization Algorithm} to achieve high performance.

Another key point of this rule is the approximation of the derivative of Dirac functions. The process for which each
occurrence of the derivative of the spiking nonlinearity is replaced by the derivative of a smooth function is called \textbf{Surrogate Gradient} ~\cite{surrogate_model}.  

\subsection{The Loihi Neuromorphic Research Chip} \label{Loihi_section}

\begin{figure}[h]
    \centering
    \includegraphics[scale=0.5]{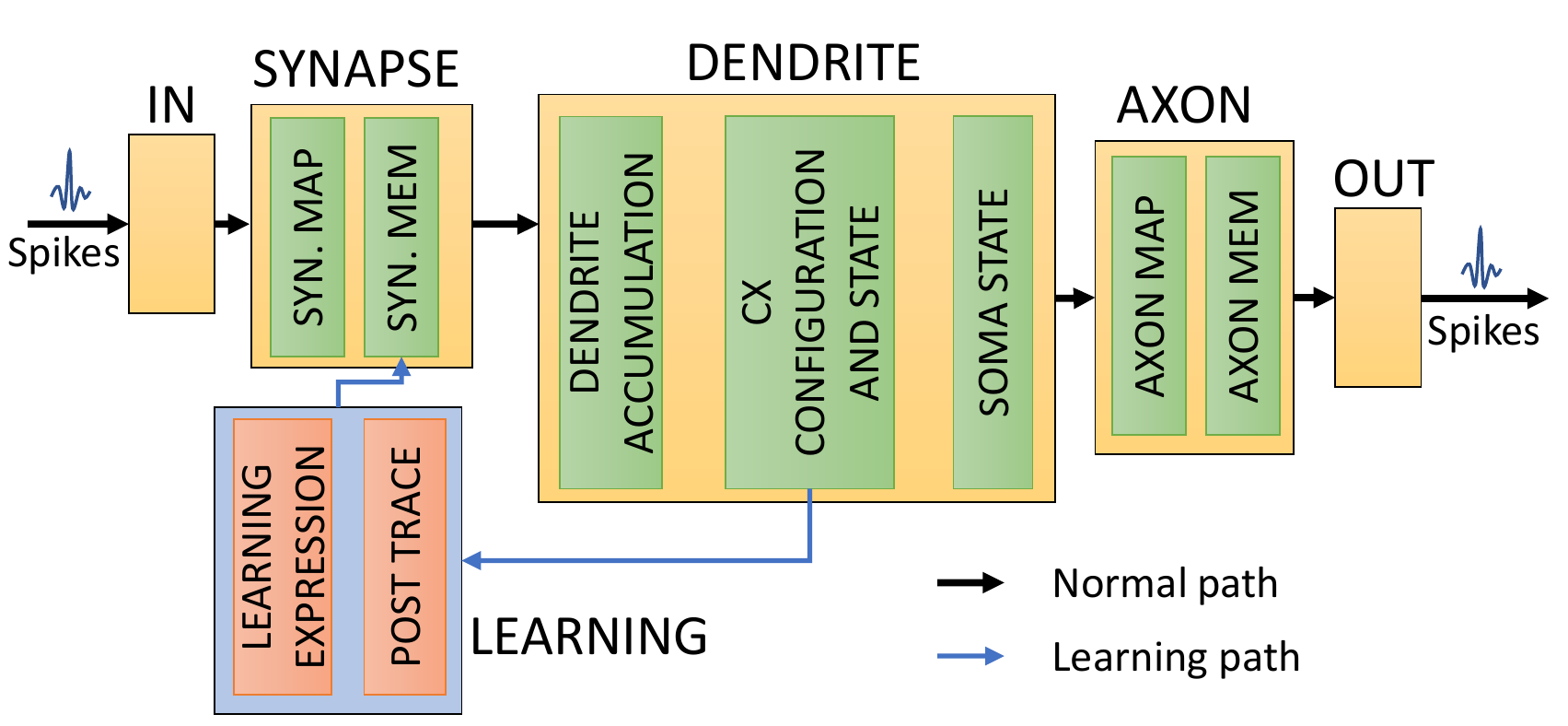}
    \caption{Simplified Neurocore abstraction mechanism on the Loihi Neuromorphic chip~\cite{Loihi_chip}.}
    \label{fig:Neuromorphic_process}
\end{figure}

Towards high energy-efficiency, it is convenient to implement SNNs on a specialized hardware, called \textbf{Neuromorphic Chip}, to guarantee high efficiency both in terms of working time and power consumption of the application. More specifically, neuromorphic hardware platforms simulate the processes that happens in the brain with one neural model, for example the LIF, using an asynchronous mechanism, as shown in Figure~\ref{fig:Neuromorphic_process}, in which every part represents one neuron attribute, for example \textit{Axon}, \textit{Synapse} and \textit{Dendrite}. In some cases, on the Chip there is also a Learning part that can be used for \textit{Online Learning} or \textit{Continual Lifelong Learning}~\cite{lifelong_l}.

There exist several neuromorphic chips developed by premier industries and academia, like $IBM\ Truenorth$~\cite{IBM_truenorth}, $SpiNNaker$~\cite{Spinnaker}, $Intel\ Loihi$~\cite{Loihi_chip}. 
The Loihi chip, which is used in this work, adopts the \textbf{CUrrent BAsed (CUBA) LIF} to model the neurons' behavior. I.e., all the neurons are a reservoir of charge (\textbf{Dendrite}), and when this overcomes the voltage threshold, there is a current spike on the output axons. This mechanism is very similar to the behavior that happens on the LIF model and it can be visualized with the help of the following Equation~\ref{eq:CUBA_LIF}.

\begin{equation}
    \dot{v}_{i}(t)=-\frac{1}{\tau_{v}} v_{i}(t)+u_{i}(t)-\theta_{i} \sigma_{i}(t),
    \label{eq:CUBA_LIF}
\end{equation}
where:
\begin{itemize}[leftmargin=*]
    \item $\tau_v$ is the leakage contribution;
    \item $u_i$ is the input current from the synapses;
    \item $v_i$ is the dendrite potential;
    \item $\theta_{i}$ is the voltage threshold;
    \item $ \sigma_{i}$ represents  the generation of an output spike on axons.
\end{itemize}

The Loihi chip is composed of \textit{neurocores} that represent groups of neurons, whose behavior is simulated by the \textit{compartments}. Every neurocore exchanges information (spikes) to the others by an asynchronous Network-on-Chip (NoC) in the form of packetized messages. To spread the spikes in an asynchronous way, a \textit{mesh operation} is used for each time step, which can be summarized in four points:

\begin{enumerate}[leftmargin=*]
    \item each neurocore independently iterates its compartments and, if a compartment is in spike firing state, this information is send with a message onto the NoC;
    \item the messages are sent to all destination neurocores;
    \item when a neurocore ends its internal distribution, it sends a barrier signal to the neighbors;
    \item when all the neurocores receive such signal, the time step is incremented.
\end{enumerate}

Every chip can contain 128 neurocores, but to implement wider and deeper SNNs, many chips can work together without any increase in the latency of message exchange.
The programmer has to follow some constraints in the implementation:

\begin{itemize}[leftmargin=*]
    \item every neurocore can have a maximum of 1024 compartments; 
    \item the max fan-in of every neurocore is 4096 pre-synapses;
    \item the max fan-out of every neurocore is 4096 post-synapses;
    \item the total synaptic fan-in state mapped to any neurocore must not exceed 128 KB.
\end{itemize}

The \textit{On-chip Learning} engine can operate to implement \textit{Online Learning} strategies or unsupervised learning with local information. On the other hand, for the back-propagation methods, the given SNN can be trained offline and then transported to the Loihi with the $NxSDK\ API$ commands. Its API gives many facilitation to the programmer. For example, the main SNN temporal parameters (synaptic, axon and refractory delay) can be configured and adapted to have a polysynchronous dynamic. Moreover, a noise injector can be activated to limit overfitting when the learning engine is used.
 
\section{Problem Analysis and General Design Decisions}\label{section:analysis}

In the classification problem that we face, we can use a supervised learning method and train the network based on the desired behavior. 
Every sample is represented by a stream of events, where a stream represents the same object to classify. In the same sample, the present spikes are correlated in time and space with the past and future spikes~\cite{N-cars_dataset}. To achieve good performance, we have to take into account this temporal correlation and use a learning method capable to exploit this property. As claimed in~\cite{STBP}, the STBP is one of the best offline learning methods, and achieves very high classification accuracy in tasks that involve event-based camera streams. It also uses both TD and SD to calculate the gradients and train the SNN. Therefore, we employ this learning method in our experiments.

This is also a real-time problem, as the system should be very reactive and perform the correct prediction in few milliseconds.
%
%
Since we want a very reactive prediction, we can use only a subset of input information, and therefore implement the \textbf{Attention Window} strategy. To find the area which focuses the attention on input data, we analyze and evaluate the event occurrences, both in train and test sets of the N-CARS dataset~\cite{N-cars_dataset}. Due to the relatively large dimension of this dataset, this study resembles with a good approximation the real problem and does not affect the generalization property of our system. 

\begin{figure}[h]
    \centering
    \includegraphics[scale=0.25]{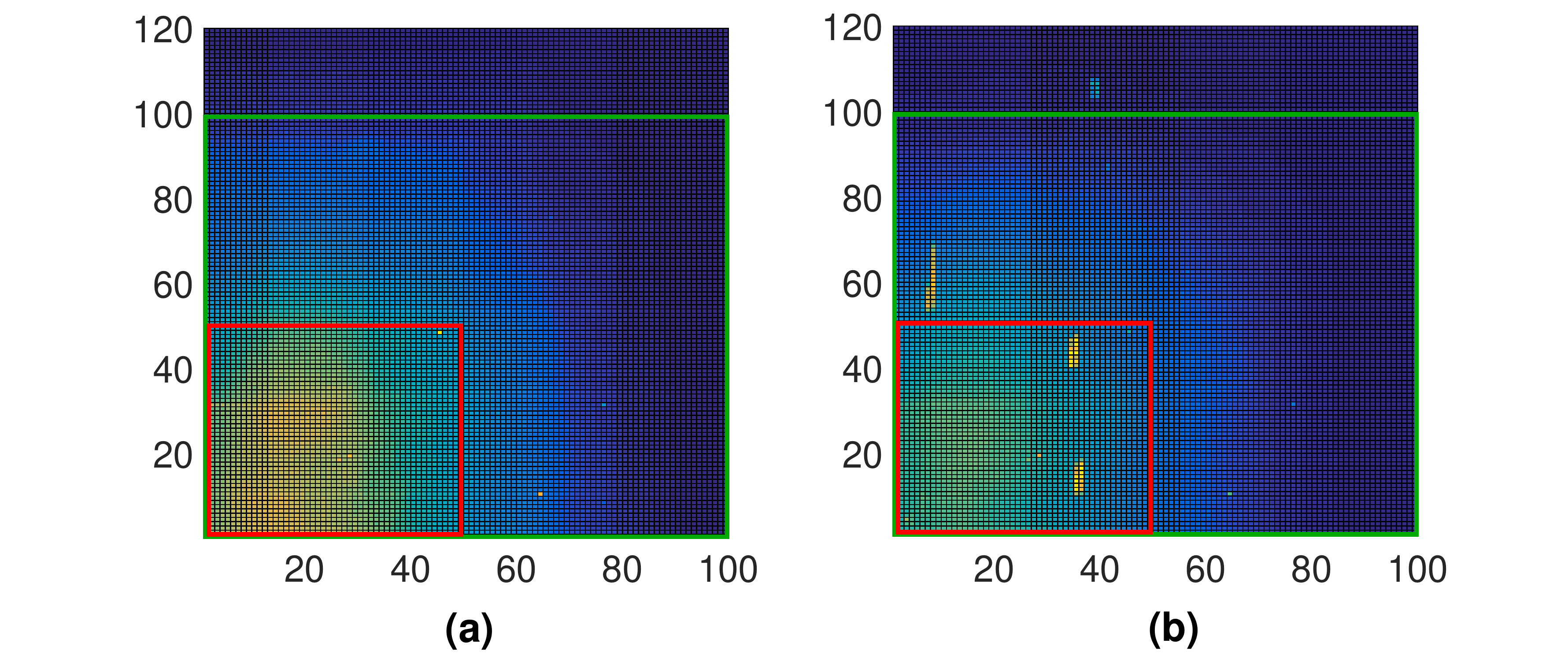}
    \caption{Event occurrences on (a) test and (b) train sets of the N-CARS dataset.}
    \label{fig:occ_N_cars_sp}
\end{figure}

The evaluation of the event occurrences in different attention windows is shown in Figure~\ref{fig:occ_N_cars_sp}. Most of the information is contained in the area of size 50$\times$50 in the bottom-left corner, both in train and test set. Hence, as reported in Table~\ref{tab:attention_window}, we can use this as the $first\ attention\ window$. The $second\ attention\ window$ has a doubled size (i.e., 100$\times$100) and also starts from the bottom-left corner.

\begin{table}[h]
    \centering
    \vspace*{10pt}
    \caption{Delimited points for \textbf{Attention windows}.}
    \label{tab:attention_window}
    \resizebox{\columnwidth}{!}{
    \begin{tabular}{c|c|c|c|c}
    Attention window    & P. 0 (x,y) & P. 1 (x,y) & P. 2 (x,y) & P. 3 (x,y)  \\
         \hline
    \textcolor{red}{First\ attention\ window}    & (0,0) &  (0,50) &  (50,50) & (50,0)\\
    \textcolor{mygreen}{Second\ attention\ window}    & (0,0) &  (0,100) &  (100,100) & (100,0)
    \end{tabular}
    }
    
\end{table}

Considering its practical implementation on an existing \textbf{Neuromorphic Hardware}, the \textbf{Intel Loihi Research Chip}, the network is designed following all the constraints of this chip, summarized in Table~\ref{tab:constrain_Loihi}.

\begin{table}[h]
    \centering
    \caption{Main constraints for developing the SNN implemented on the \textbf{Intel Loihi Neuromorphic Research chip}.}
    \begin{tabular}{c|c}
         Property & Constrain\\
         \hline
         Maximum Compartments per Core & 1024 Compartments\\
         Maximum fan-in of a Core & 4096 Pre-Synapses\\
         Maximum fan-out of a Core & 4096 Post-Synapses\\
         Synaptic fan-in state size & 128 KB
    \end{tabular}
    
    \label{tab:constrain_Loihi}
\end{table}

A summary of the general decisions taken after analyzing the problem is shown in Table~\ref{tab:decision_after_analysis}.

\begin{table}[h]
    \centering
    \caption{General decisions taken after analyzing the problem.}
    \begin{tabular}{c|c}
         \textbf{Properties of the problem} & \textbf{Decision}\\
         \hline
         Knowledge of the correct output & Use supervised learning rule\\
         Time and space correlation & Take into account TD and SD\\
         Real-time & Use simplest SNN\\
         High performance of vision sensor & Use event-based camera\\
         Good profiling of real problem & Use N-CARS dataset\\
         Many information in limited area & Use attention windows \\
         Low power consumption & Use Neuromorphic Chip \\
         
    \end{tabular}
    
    \label{tab:decision_after_analysis}
\end{table}

\section{CarSNN: Our Proposed SNN for event-based Cars vs. Background Classification}\label{section:Methodology}

\begin{figure}[h]
    \centering
    \includegraphics[width=\columnwidth]{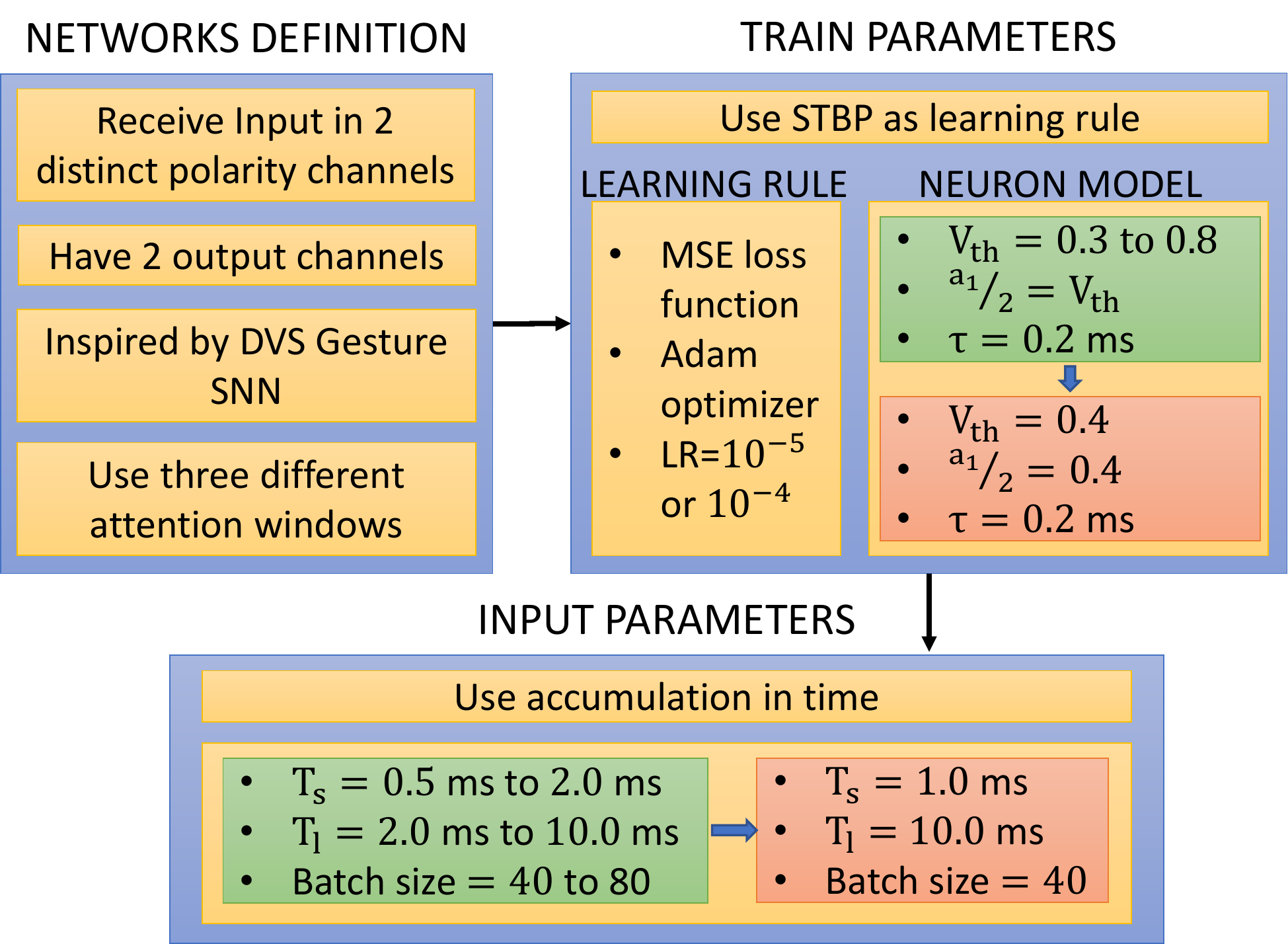}
    \caption{Three-step process followed to design our \textit{CarSNN} with the training and feeding input parameters. }
    \label{fig:parameter_definition_process}
\end{figure}

Our methodology to design the SNN model for the \textit{``cars vs. background''} classification, which we call \textbf{CarSNN}, is composed as a three-step process, as shown in Figure~\ref{fig:parameter_definition_process}. After the definition of the SNN model architectures considering different attention windows in Section~\ref{subsection:model_design}, the methods for finding the parameters for SNN training and feeding input data are discussed in Sections~\ref{subsection:method_training} and~\ref{subsection:method_input_data}, respectively.

\subsection{CarSNN Model Design}\label{subsection:model_design}

To achieve good classification results, our \textit{CarSNN} receives the input events in two distinct polarity channels, one for positive and one for negative events. Towards the generalization of the problem, we consider this as a multi-classification problem (i.e., not as a simple binary classification problem). Therefore, the output layer of the \textit{CarSNN} consists of two neurons that correspond to the two possible classes, one for cars and the other for background objects.

Since the architecture proposed in~\cite{Motivational_case} achieved high classification accuracy and low latency on the IBM DVS128 gesture dataset~\cite{GESTURE_DATA}, we modify this model to correctly function for the N-CARS dataset. Compared to the model of~\cite{Motivational_case}, our \textit{CarSNN} has different output channels, kernel size and padding on the first convolutional layer, and different sizes of the last two dense layer. 

Based on the attention window analysis, we develop three different SNNs for the three different sizes of input images:
\begin{enumerate}[leftmargin=*]
    \item Size $128 \times 128$ (Table~\ref{tab:network_128_128}): the model is very similar to the SNN proposed in~\cite{Motivational_case}. Since this size overcomes the N-CARS dataset image size, which is $120 \times 100$, the exceeded pixels do not produce spikes and are padding by zeros (no event). This image size is equal to the resolution of one of the most used DVS camera~\cite{ATIS_camera}. Therefore, this network can be easily implemented with it.
    \item Size $50 \times 50$ (Table~\ref{tab:network_50_50}): this uses the first attention window as described in Section~\ref{section:analysis}.
    \item Size $100 \times 100$ (Table~\ref{tab:network_100_100}): this uses the second attention window as described in Section~\ref{section:analysis}.
    
\end{enumerate}

\begin{table}[h]
    \centering
    \caption{SNN model for full-size images (input size $128 \times 128$).}
    \begin{tabular}{|c|c|c|c|c|c|}
        \hline Layer type& In ch. & Out ch. & Kernel size & Padding & Stride \\
        \hline
        Av. pooling & 2 & 2 & 4 & $-$ & $-$ \\
        Convolution & 2 & 32 & 3 & 1 & 1 \\
        Av. pooling & 32 & 32 & 2 & $-$ & $-$ \\
        Convolution & 32 & 32 & 3 & 1 & 1 \\
        Av. pooling & 32 & 32 & 2 & $-$ & $-$ \\
        Dense & 2048 & 1024 & $-$ & $-$ & $-$ \\
        Dense & 1024 & 2 & $-$ & $-$ & $-$ \\
        \hline
    \end{tabular}
    \label{tab:network_128_128}
\end{table}

\begin{table}[h]
    \centering
    \caption{SNN model for first attention window (input size $50 \times 50$).}
    \begin{tabular}{|c|c|c|c|c|c|}
        \hline Layer type & In ch. & Out ch. & Kernel size & Padding & Stride \\
        \hline
        Av. pooling & 2 & 2 & 4 & $-$ & $-$ \\
        Convolution & 2 & 32 & 3 & 1 & 1 \\
        Av. pooling & 32 & 32 & 2 & $-$ & $-$ \\
        Convolution & 32 & 32 & 3 & 1 & 1 \\
        Av. pooling & 32 & 32 & 2 & $-$ & $-$ \\
        Dense & 512 & 144 & $-$ & $-$ & $-$ \\
        Dense & 144 & 2 & $-$ & $-$ & $-$ \\
        \hline
    \end{tabular}
    \label{tab:network_50_50}
\end{table}

\begin{table}[h]
    \centering
    \caption{SNN model for second attention window (input size $100 \times 100$).}
    \begin{tabular}{|c|c|c|c|c|c|}
        \hline Layer type & In ch. & Out ch. & Kernel size & Padding & Stride \\
        \hline
        Av. pooling & 2 & 2 & 4 & $-$ & $-$ \\
        Convolution & 2 & 32 & 3 & 1 & 1 \\
        Av. pooling & 32 & 32 & 2 & $-$ & $-$ \\
        Convolution & 32 & 32 & 3 & 1 & 1 \\
        Av. pooling & 32 & 32 & 2 & $-$ & $-$ \\
        Dense & 1568 & 512 & $-$ & $-$ & $-$ \\
        Dense & 512 & 2 & $-$ & $-$ & $-$ \\
        \hline
    \end{tabular}
    \label{tab:network_100_100}
\end{table}

\subsection{Parameters for Training} \label{subsection:method_training}

Using a supervised learning rule based on backpropagation, like the STBP, it is possible to tune several hyper-parameters.

We focus our attention on:
\begin{itemize}[leftmargin=*]
    \item \textbf{loss function}: we adopt the \textbf{Mean Squared Error} (\textbf{MSE}) loss criterion, since it achieves the highest performance in~\cite{STBP};
    
    \item \textbf{optimizer}: we use \textbf{Adam}~\cite{adam}, because it seems the best for the STBP;
    
    \item \textbf{learning rate} (\textbf{lr}): after some preliminary tests, we find the best value is around \textbf{$1e^{-5}$} and \textbf{$1e^{-4}$}, where with the latter value the training is faster and the SNN achieves good accuracy results in fewer epochs. 
    
\end{itemize}
Since the adopted learning rule is directly implemented on the SNNs with LIF neurons, other specific parameters can be adjusted. The kernel of the LIF neuron model can be described by Equation~\ref{eq:system_STBP_model} (discussed in section ~\ref{subsection:STBP}).
We focus on the formalization of the membrane potential update ($u_{i}^{t+1, n}$) and highlight the membrane potential decay factor $\tau$ (Equation~\ref{eq:STBP_model_one_formula}).

\begin{equation}
    u_{i}^{t+1, n} =u_{i}^{t, n} \tau (1-o_{i}^{t,n})+\sum_{j=1}^{l(n-1)} w_{i j}^{n} o_{j}^{t+1, n-1}+b_{i}^{n}
    \label{eq:STBP_model_one_formula}
\end{equation}

Another fundamental parameter of a LIF neuron is its \textbf{threshold} (\textbf{$V_{th}$}). If the membrane potential overcomes this value an output spike is generated and the potential is reset to a specific value. For each experiment, all the neurons have the same $V_{th}$ and $0$ as reset value. 

The third parameter that needs to be set (\textbf{$\frac{a_{1}}{2}$}) is related to the approximation of the derivative of spiking nonlinearity. We use the rectangular pulse function defined in Equation~\ref{eq:STBP_pulse_fcn}:

\begin{equation}
   h_{1}(u)=\frac{1}{a_{1}} \operatorname{sign}\left(\left|u-V_{t h}\right|<\frac{a_{1}}{2}\right)
   \label{eq:STBP_pulse_fcn}
\end{equation}

In the following, we perform some experiments to set the previously-discussed parameters, with a particular focus on \textbf{$V_{th}$}. We made these decisions:
\begin{itemize}[leftmargin=*]
    \item \textbf{$V_{th}$}: we change this value from 0.3 to 0.8 and evaluate which curve achieves the best accuracy;
    \item \textbf{$\frac{a_{1}}{2}$}: it assumes the same value of the threshold, as this assumption is made in~\cite{STBP};
    \item \textbf{$\tau$}: this value must be small to have good approximation of the neuron model and in particular of $f(o_{i}^{t, n})$ (see Equation~\ref{eq:system_STBP_model_f}). We set it to be equal to $0.2\ ms$.
\end{itemize}

\begin{figure}[h]
    \centering
    \includegraphics[width=\linewidth]{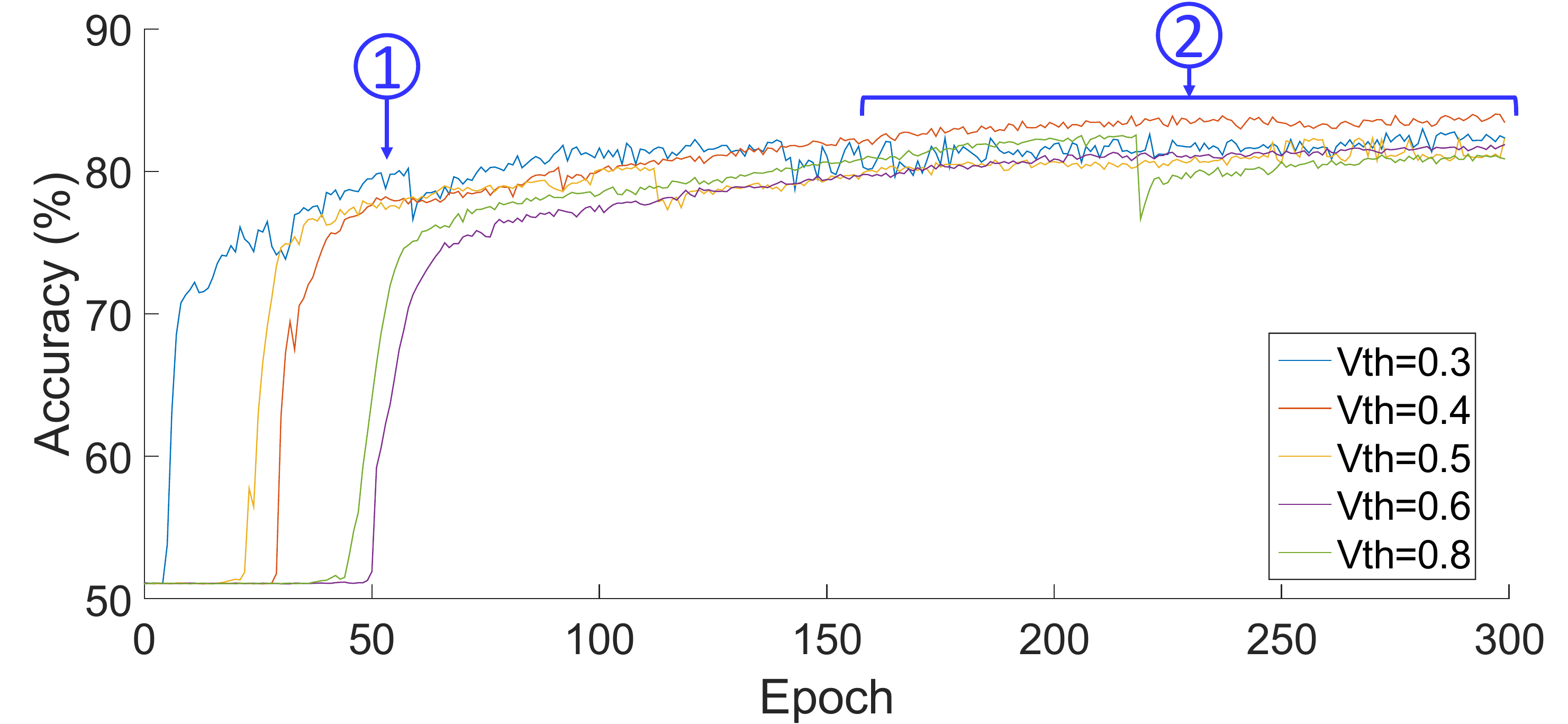}
    \caption{Percentage of accuracy for the experiment made to evaluate the best value for \textbf{$V_{th}$}.}
    \label{fig:tune_th_train}
\end{figure}

To speed up this process and have good performance, we introduce an accumulation mechanism. We accumulate the spikes at a constant time-rate called sample time ($T_{sample}$); for these first experiments this value is set to $10\ ms$. Every $T_{sample}$ time we construct a new input image that feeds the SNN. The events that compose the image are summed by the following simple rule, based on which each pixel can have a maximum of one spike per channel.  Each derived image is maintained stable to the input of the proposed SNN by a time window of 15 time steps. Therefore, this accumulation mode can compress the input information. The accuracy that we evaluate is referred to every single sample (i.e., accumulated image) on a training of 300 epochs. Table~\ref{tab:tune_th_train} and Figure~\ref{fig:tune_th_train} report the results of these experiments, where we use the SNN with the full size image (Table~\ref{tab:network_128_128}).

\begin{table}[h]
    \centering
    \caption{Experiments to find the best value of $V_{th}$.}
    \resizebox{\columnwidth}{!}{
    \begin{tabular}{c|c|c|c|c|c|c|c}
        input size & $V_{th}$ & $\frac{a_{1}}{2}$ & $\tau$ & $T_{sample}$ & batch size & lr & accuracy\\ 
        \hline
        & & & ms & ms &  &  & \% \\ 
        \hline
        $128 \times 128$ & 0.3 & 0.3 & 0.2 & 10  & 20 & $1e^{-5}$ & 83.0 \\
        $128 \times 128$ & \textbf{0.4} & 0.4 & 0.2  & 10  & 20 & $1e^{-5}$ & \textbf{84.0} \\
        $128 \times 128$ & 0.5 & 0.5 & 0.2  & 10  & 20 & $1e^{-5}$ & 82.4 \\
        $128 \times 128$ & 0.6 & 0.6 & 0.2  & 10  & 20 & $1e^{-5}$ & 81.9 \\
        $128 \times 128$ & 0.8 & 0.8 & 0.2  & 10  & 20 & $1e^{-5}$ & 82.6 \\
    \end{tabular}
    }
    \label{tab:tune_th_train}
\end{table}

From Table~\ref{tab:tune_th_train}, we notice that the best accuracy is achieved when \textbf{$V_{th}$} is equal to \textbf{0.4}. Moreover from Figure~\ref{fig:tune_th_train}, we can notice that, while a $V_{th}$ equal to 0.3 leads to a relatively high accuracy after a few epochs (see pointer \rpoint{1} in Figure~\ref{fig:tune_th_train}), the training curve with $V_{th}$ equal to 0.4 have less instability than for the other experiments (see pointer \rpoint{2}). These two reasons lead us to choose \textbf{0.4} for the \textbf{$V_{th}$} parameter.

\subsection{Parameters for Feeding the Input Data} \label{subsection:method_input_data}


As discussed in the previous section, the input spikes are given to the SNN with an accumulation strategy, to speedup the training. From the experiments conducted in Table~\ref{tab:tune_th_train}, despite this limitation, we notice that the accuracy is quite high. Therefore, we keep this property that gives us some advantages:

\begin{itemize}[leftmargin=*]
    \item decrease power consumption;
    \item increase the reactivity of the system, because input data are compressed.
\end{itemize}

Moreover, We also give an upper bound to the latency of the system of $10\ ms$. Hence, for the train, we take only $10\ ms$ from the dataset sample stream with a random initial point. This is defined as the maximum acceptable sample length ($T_l$). With this constraint, two different approaches can be adopted:

\begin{enumerate}
    \item accumulate the spikes every $T_l$ time ($T_{sample} = T_l$) and do the prediction on a unique input image for the entire input stream, as we did in the previous experiments;
    \item accumulate the spikes in order to have more than one input image for every input stream ($T_{sample} < T_l$), then see what is the class with majority prediction.
\end{enumerate}

We conduct some analyses to find the best sample time and the variation of the accuracy with two different batch sizes (BS on Table~\ref{tab:tune_dataset_train}). In these experiments we use the second approach for the image accumulation and we set the parameters as follows: $V_{th}=0.4$, $\frac{a_{1}}{2}=0.4$, $\tau=0.2\ ms$.


The training lasts for 200 epochs and to speed up this process, as discussed in section~\ref{subsection:method_training}, we use a learning rate equal to $1e^{-4}$ and a minimum batch size of 40. We also use three different metrics to evaluate the accuracy:

\begin{itemize}[leftmargin=*]
    \item one shot accuracy on test data ($acc._{s}$): it is the accuracy found on all the samples taken at $T_{s}$ of the test dataset;
    \item accuracy on test data ($acc._{test}$): it is the accuracy for all the sample stream of the test dataset, computed based on the majority prediction of the part of the stream with sample length equal to $T_l$;  
    \item accuracy on train data ($acc._{train}$): it is the counterpart of the accuracy on test data, but calculated on train streams of the dataset.
\end{itemize}

\begin{table}[h]
    \centering
    \caption{Experiments to find the best value for $T_s$, $T_l$ and batch size.}
    \resizebox{\columnwidth}{!}{
    \begin{tabular}{c|c|c|c|c|c|c|c}
        Input size & $T_{s}$ & $T_{l}$ & BS & lr & $acc._{s}$ & $acc._{test}$ & $acc._{train}$\\ 
        \hline
         & ms & ms &  &  & \% & \% & \%\\ 
        \hline
        $128 \times 128$ & 1.0 & 2.0 & 80  & $1e^{-4}$ & 80 & 79 & 83 \\
        $128 \times 128$ & 1.0 & 4.0 & 80  & $1e^{-4}$ & 80 & 80 & 86 \\
        $128 \times 128$ & 1.0 & 6.0 & 80  & $1e^{-4}$ & 51 & 51 & 51 \\
        $128 \times 128$ & 1.0 & 8.0 & 80  & $1e^{-4}$ & 80 & 79 & 89 \\
        $128 \times 128$ & 1.0 & 2.0 & 40  & $1e^{-4}$ & 80 & 77 & 86 \\
        $128 \times 128$ & 1.0 & 4.0 & 40  & $1e^{-4}$ & 80 & 83 & 88 \\
        $128 \times 128$ & 1.0 & 6.0 & 40  & $1e^{-4}$ & 72 & 70 & 90 \\
        \textbf{$128 \times 128$} & \textbf{1.0} & \textbf{8.0} & \textbf{40}  & \textbf{$1e^{-4}$} & \textbf{81} & \textbf{86} & 91 \\
        $128 \times 128$ & 1.0 & 10.0 & 40  & $1e^{-4}$ & 80 & 86 & \textbf{94} \\
        $128 \times 128$ & 2.0 & 10.0 & 40  & $1e^{-4}$ & 51 & 51 & 51 \\
        $100 \times 100$ & 0.5 & 10.0 & 40  & $1e^{-4}$ & 75 & 80 & 84 \\
        \textbf{$100 \times 100$} & \textbf{1.0} & \textbf{10.0} & \textbf{40}  & \textbf{$1e^{-4}$} & \textbf{81} & \textbf{85} & \textbf{92} \\
        $100 \times 100$ & 2.0 & 10.0 & 40  & $1e^{-4}$ & 51 & 51 & 51 \\
        $50 \times 50$ & 0.5 & 10.0 & 40  & $1e^{-4}$ & 67 & 71 & 79 \\
        $50 \times 50$ & 1.0 & 10.0 & 40  & $1e^{-4}$ & 71 & 75 & 81 \\
        \textbf{$50 \times 50$} & \textbf{2.0} & \textbf{10.0} & \textbf{40}  & \textbf{$1e^{-4}$} & \textbf{74} & \textbf{77} & \textbf{83} \\

    \end{tabular}
    }    
    \label{tab:tune_dataset_train}
\end{table}


The results in Table~\ref{tab:tune_dataset_train} provide us the necessary feedback for setting the value of $T_s$. If it is small (i.e., $0.5\ ms$) there are more points for the same stream sample. However, it is very difficult to train the SNNs, because the accumulation has not effect and the temporal correlation is lost. On the other hand, the accuracy is low when we use high $T_s$ (i.e., $2\ ms$). The best trade-off is obtained when $T_s$ is equal to $1\ ms$.

Moreover, the batch size influences the training process. Indeed, to have high accuracy, the value of BS should be limited to 40.  

In the first experiments of Table~\ref{tab:tune_dataset_train}, we consider only the variation of $T_l$ and BS. With constant BS and same value of $acc._s$, the $acc._{test}$, as expected, increases or remains stable with the increasing of $T_l$. This behavior is due to having more sub-predictions to compute the final result when $T_l$ is large. The changes in the $acc._s$ are justified by the non-deterministic training process.

\section{Evaluation of our CarSNN Implemented onto the Loihi Neuromorphic Chip}\label{section:result}

\begin{figure}[h]
	\centering
	\includegraphics[width=1\linewidth]{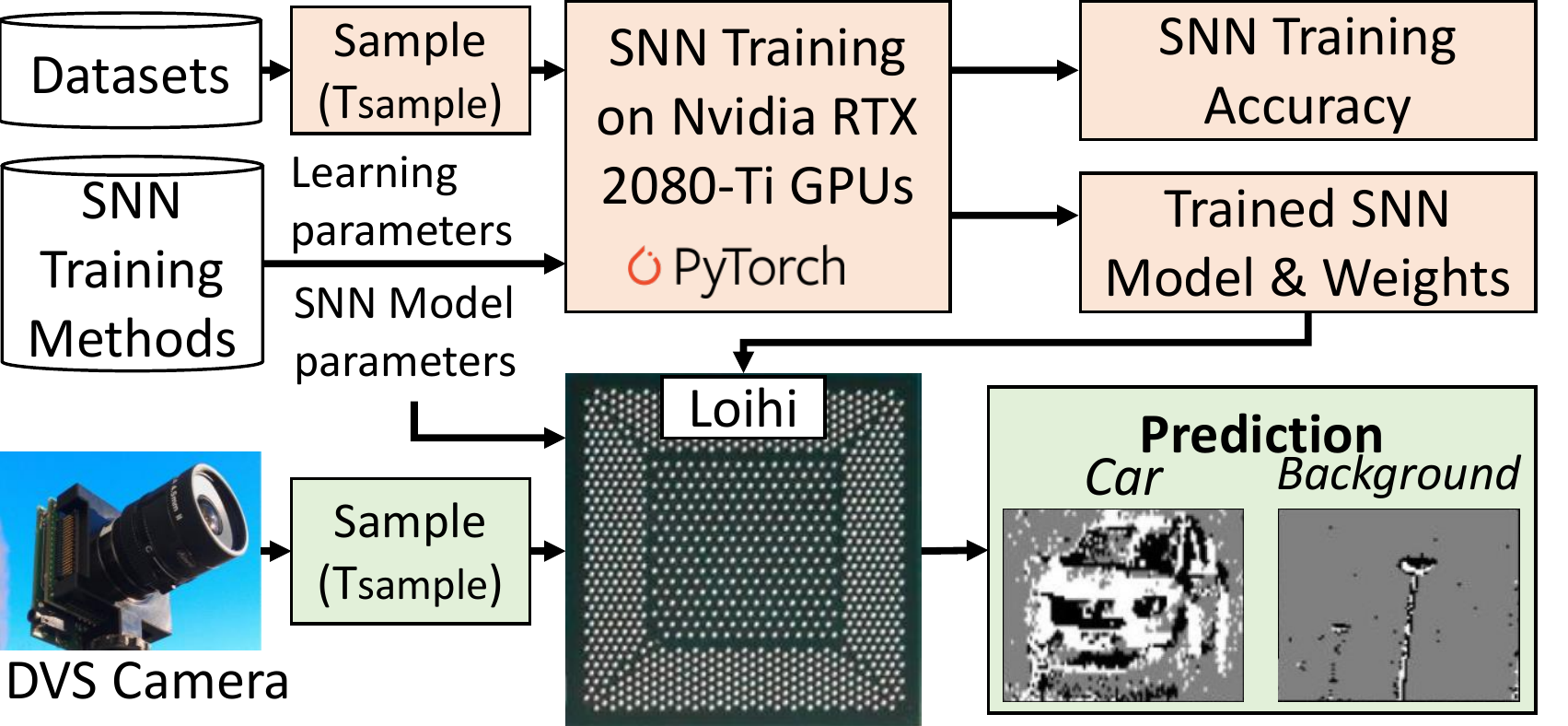}
	\caption{Setup and tool-flow for conducting our experiments.}
	\label{exp_setup}
\end{figure}

The STBP learning method is based on the backpropagation, without using the local information. Moreover, the gradients are computed with Equations~\ref{eq:STBP_diff_eq_end_1} and~\ref{eq:STBP_diff_eq_end_2}, which are not directly implementable into the on-chip learning section of the Intel Loihi Neuromorphic hardware. For these reasons, our \textit{CarSNN} is trained offline and then the resulting parameters are mapped onto the neuromorphic chip. An overview of the tool-flow for conducting the experiments is shown in Figure~\ref{exp_setup}.

\subsection{Experimental Setup and Accuracy Results for CarSNN Offline Training}\label{subsection:exp_setup}

Coherently with the analysis due in previous sections, in order to train and validate the prediction system we use the N-CARS dataset. We take into account this dataset also for the two fundamental reasons that it collects event-based camera streams and it is the largest labeled event-based dataset acquired in real-world conditions~\cite{N-cars_dataset}.

We describe the SNNs using the PyTorch library~\cite{PyTorch}. In these codes, we model the SNNs' functional behavior with the implementation of the Equation~\ref{eq:STBP_model_one_formula} that contains the mechanism to update the membrane potential. 

We run the experiments on a workstation having CentOS Linux release 7.9.2009 as operating system and equipped with an Intel Core i9-9900X CPU and Nvidia RTX 2080-Ti GPUs. 

The setting of the hyper-parameters follows the analyses made in sections ~\ref{section:analysis} and ~\ref{section:Methodology}, and are summarized in Table~\ref{tab:results_offline_parameter}. 

\begin{table}[h]
    \centering
    \caption{Parameters of the experiments.}
    \resizebox{\columnwidth}{!}{
    \begin{tabular}{c|c|c|c|c|c|c|c}
         Epochs & $T_{s}$ & $T_{l}$ & BS & lr & $V_{th}$ & $\frac{a_{1}}{2}$ & $\tau$\\ 
        \hline
         & ms & ms &  &  &  &  & ms\\ 
        \hline
    200 & 1.0 & 10.0 & 40  & $1e^{-3}\ to\ 1e^{-6}$ & 0.4 & 0.4 & 0.2 \\
    \end{tabular}
    }    
    \label{tab:results_offline_parameter}
\end{table}

The dataset streams are randomly shuffled and the sample of $T_l$ is taken starting from a random initial point. We set the BS to 40, that gives best accuracy in the previous experiments (according to Table~\ref{tab:tune_dataset_train}), and maintains a reasonable the training time duration. We set the same values of $T_s=1\ ms$ and $T_l=10\ ms$ for the three experiments, to have a fair comparison between them. These two values leverage the trade-off found from the results in Table~\ref{tab:tune_dataset_train}. The parameters for the SNN model are the same used in Section~\ref{subsection:method_input_data}. The learning rate ($lr$) decreases by 0.5 every 20 epochs, starting for the value $1e^{-3}$. With this approach, the accuracy slightly increases, compared to having a fixed $lr$.

To ease the model mapping onto the Loihi Neuromorphic Chip, only the weights are updated during training, while the bias is forced to 0. The train lasts for 200 epochs and every sample taken at $T_s$ time is evaluated for 20 time steps. With these hardware and software settings, the training for one single epoch on all the dataset samples is measured to be about 300 seconds.
For the inference, the mean latency for all samples, given at the time $T_s$, is about 0.8 ms. 
Table~\ref{tab:results_offline} shows the results in terms of the same accuracy policies as defined in Section~\ref{subsection:method_input_data}.

\begin{table}[h]
    \centering
    \caption{Results of the offline training experiments.}
    \begin{tabular}{c|c|c|c}
        Input size & $acc._{s}$ & $acc._{test}$ & $acc._{train}$\\ 
        \hline
        & \% & \% & \%\\ 
        \hline
        $128 \times 128$ & 80.1 & 85.7 & 93.6 \\
        $100 \times 100$ & 80.5 & 86.3 & 95.0 \\
        $50 \times 50$ & 72.6 & 78.7 & 85.3 \\
        
    \end{tabular}
    \label{tab:results_offline}
\end{table}

The accuracy values for the attention window of size $100 \times 100$ are comparable to the results for the full image size ($128 \times 128$), and indeed exhibit slightly higher $acc._{test}$ and $acc._{train}$. It can be explained because the cropped part of the sample is not important for the correct classification and might lead to an SNN misfunctioning. On the other hand, the input values consisting of a small part of the original image ($50 \times 50$) lead to a significant accuracy decrease.

Moreover, from the results in Table~\ref{tab:results_offline}, we can notice an overfitting, due to the gap between $acc._{test}$ and $acc._{train}$, which can be considered to be the upper bound of the accuracy for our developed \textit{CarSNN} models.


\subsection{CarSNN Implemented on Loihi}\label{subsection:loihi_result}

To implement our network on the Intel Loihi Neuromorphic Chip we have to exploit some similarity between its model and our offline model used for the previous experiments. Equation~\ref{eq:comp_voltage} reports how the Compartment Voltage ($CompV$), which represents the membrane voltage of a neuron, is evaluated by the neuromorphic hardware~\cite{Loihi_chip}.
\begin{equation}
    CompV_{t+1} =CompV_{t} \frac{2^{12} - \delta_v}{2^{12}}+CompI_{t+1}+bias
    \label{eq:comp_voltage}
\end{equation}

The Compartment Current ($CompI$) is formulated by Equation~\ref{eq:comp_current}, where the sum expression represents the accumulation of the weighted incoming spikes from $j^{th}$ pre-synaptic neuron.

\vspace*{-10pt}
\begin{equation}
\resizebox{.89\columnwidth}{!}{
    $CompI_{t+1}=CompI_{t} \frac{2^{12} - \delta_i}{2^{12}} +         2^{6+wgtExp}\sum_jw_{j} s_{j_{t+1}}$
    }
    \label{eq:comp_current}
\end{equation}

In Equations~\ref{eq:comp_voltage} and~\ref{eq:comp_current}, we can set the following parameters:

\begin{itemize}[leftmargin=*]
    \item $\delta_i$: Compartment Current Decay;
    \item $\delta_v$: Compartment Voltage Decay;
    \item $bias$: bias component on $CompV$;
    \item $wgtExp$: value used to implement very different weights between different SNN layers.
\end{itemize}

Comparing the formulation of our offline model (i.e., Equation~\ref{eq:STBP_model_one_formula}) and the Equation~\ref{eq:comp_voltage}, we notice its similarity to Equations from~\ref{eq:simila_compV} to~\ref{eq:simila_bias}.  
\begin{equation}
\vspace*{-6pt}
    CompV_{t}=u_{t}
    \label{eq:simila_compV}
\end{equation}

\begin{equation}
\vspace*{-6pt}
    CompI_{t} =\sum_{j} w_{j} o_{j_{t+1}}\ \ \ \ if\ \delta_i=2^{12}
    \label{eq:simila_compI}
\end{equation}

\begin{equation}
\vspace*{-12pt}
    \frac{2^{12} - \delta_v}{2^{12}}=\tau
    \label{eq:simila_tau}
\end{equation}

\begin{equation}
\vspace*{-4pt}
    bias=b
    \label{eq:simila_bias}
\end{equation}

We implement only the \textit{CarSNN} described in Table~\ref{tab:network_128_128}, which achieves good offline accuracy results (as indicated in Table~\ref{tab:results_offline}) and it represents the most complex developed network, based on latency, power consumption and number of neurons.

The Loihi Neuromorphic Hardware uses only 8 bits for the storage of weights. The maximum range of our weights is $(-7, 6)$. Since these values are very different between layers and the $wgtExp$ is limited we:

\begin{enumerate}[leftmargin=*]
    \item multiply weights and $V_{th}$ by 25 (this value do not consider the default multiplication for $2^6$ of weights and $V_{th}$ made on the Loihi);
    \item use all the 8 bits to store our values.
\end{enumerate}
    
According to Equations~\ref{eq:simila_compV}-\ref{eq:simila_bias}, the other neuromorphic hardware parameters can be adjusted.

All the setup parameters are summarized in Table~\ref{tab:loihi_param}.

\begin{table}[h]
    \centering
    \caption{Translation of parameters to the Loihi Chip.}
    \resizebox{\columnwidth}{!}{
    \begin{tabular}{c|c|c|c|c|c}
        \multicolumn{3}{c}{Offline implementation}&\multicolumn{3}{c}{Loihi implementation}
        \\
        \hline
        Parameter & Value & Precision & Parameter & Value & Precision \\ 
        \hline
        $V_{th}$ & 0.4 & Floating point 64 bits & $V_{th\ mant}$ & 10 & Fixed point 12 bits\\
        $weight$ & $\times 1$ & Floating point 64 bits & $weight$ & $\times 25$ & Fixed point 8 bits\\
        $\tau$ & 0.2 & Floating point 64 bits & $\delta_v$ & 3276 & Fixed point 12 bits\\
        $b$ & 0 & Floating point 64 bits & $bias$ & 0 & Fixed point 8 bits\\
        $-$ & $-$ & Floating point 64 bits & $\delta_i$ & 0 & Fixed point 12 bits\\
    \end{tabular}
    }    
    \label{tab:loihi_param}
\end{table}

We define our \textit{CarSNN} using the Intel Nx SDK API version 0.9.5 and run it on the Nahuku32 partition, in particular we use the NxTF Layers, such as NxConv2D, NxAveragePooling2D and NxDense utilities.
This kind of implementation is useful to automatically improve the performance of the SNN in a simple manner.
The \textit{CarSNN} is tested on the N-CARS dataset. Every sample at $T_s$ is replicated for 10 timestep and between samples we insert a blank time of 7 timestep. The number of timesteps per inference is 17. This decision is necessary to follow the real-time constraint of a maximum inference latency of 1 ms. 

In the results reported in Table~\ref{tab:result_online_Loihi}, 
the mean latency, referred to the time used to evaluate every sample at $T_s$, is calculated through the multiplication between the mean total execution time (in timesteps) and the number of timesteps per inference.

On the other hand, the maximum latency is referred to the maximum ``spiking time'' for every timestep, considering the time in which the Loihi Chip is used and makes the classification decision. This value can be used to evaluate whether the latency constraint is met. It does not include the time overhead used to exchange results between the chip and the host system, that can be suppressed by directly using output ports.  

\begin{table}[h]
    \centering
    \caption{Results of the \textit{CarSNN} implemented onto the Loihi Chip.}
    \resizebox{\columnwidth}{!}{
    \begin{tabular}{c|c|c|c|c|c|c}
        $acc._s$ & $acc._{test}$ & 
        Neurons & Synapses & Neurocores & Mean latency & Max latency \\ 
        \hline
        \% & \% & 
        number & number & number & $\mu$s & $\mu$s \\
        \hline
        72.16 & 82.99 & 
        54,274 & 5,122,048 & 151
        & 899.6 & $\approx$ 700 \\
    \end{tabular}
    }
    \label{tab:result_online_Loihi}
\end{table}

\begin{figure}[h]
    \centering
    \includegraphics[width=\linewidth]{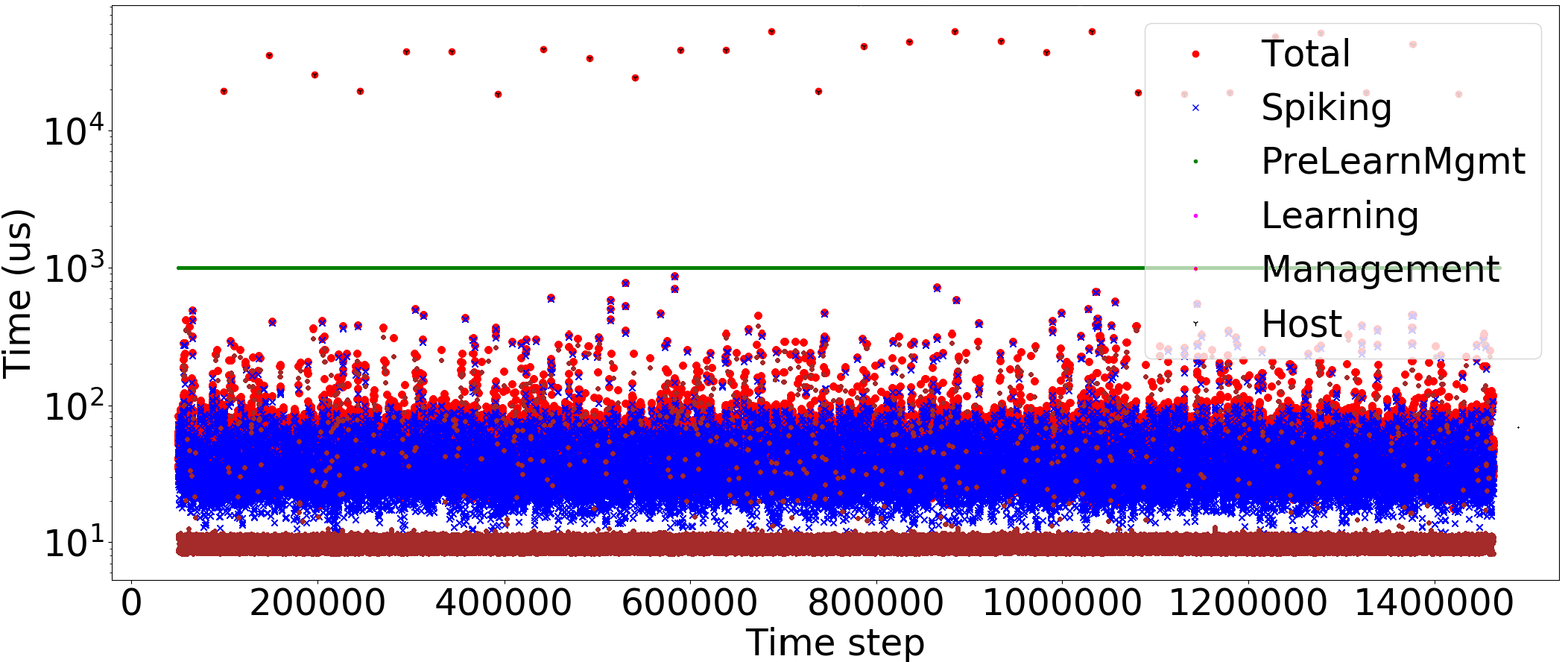}
    \caption{Execution time for every timestep. The \textcolor{mygreen_time_graph}{green line} represents the limit for the spiking time and it is set to 1 ms ($T_s$).}
    \label{fig:Graph_ex_time}
\end{figure}

From Table~\ref{tab:result_online_Loihi} and the Figure~\ref{fig:Graph_ex_time}, the following observations can be made:

\begin{itemize}[leftmargin=*]
    \item The $acc._{test}$ for the implementation onto the Loihi chip is $2.6 \%$ lower than the offline application.
    \item The maximum latency does not exceed $T_s$ ($1\ ms$).
    
\end{itemize}

Table~\ref{tab:power_and_energy_consumption_on_loihi} describes the power and energy consumption of the application implemented on the Neuromprphic Chip. In particular:

\begin{itemize}[leftmargin=*]
    \item \textit{LakeMounts Power}: it is the consumption of the embedded processors~\cite{Loihi_chip} used to menage neurons and exchange messages with the host system.
    \item \textit{Neuro-cores Power}: it represents the consumption for the neurons.
    \item \textit{System Power}: it is the consumption of the entire system, where a large part of it is represented by the static power used for the inactive chip of the used partition. It uses only 2 chips out of 32.
    \item \textit{Energy per inference}: it is the mean energy consumed to classify one sample.
\end{itemize}

\begin{table}[h]
    \centering
    \caption{Power and energy consumption of the \textit{CarSNN} implemented onto the Loihi Chip.}
    \resizebox{\columnwidth}{!}{
    \begin{tabular}{c|c|c|c}
        LakeMounts Power & Neuro-cores Power & System Power & Energy per Inference\\ 
        \hline
        mW & mW & mW & $\mu$J\\
        \hline
        40.8 & 314.5 & 1375.4 & 319.7\\
    \end{tabular}
    }
    \label{tab:power_and_energy_consumption_on_loihi}
\end{table}

Hence, Table~\ref{tab:power_and_energy_consumption_on_loihi} reports the power and energy consumption of the application implemented onto the Loihi Chip, that is several orders of magnitude lower than the same measured on GPUs.

\subsection{Comparison with the State-of-the-Art}\label{subsection:comparison}

To the best of our knowledge, \textit{CarSNN} is the the first Spiking Convolutional Neural Network (CNN) designed to perform event-based \textit{``cars vs. background''} classification on neuromorphic hardware. This is also the first method that uses statistic analysis of events occurrences to indicate different attention windows on it. In this paper, we use a simple yet efficient mechanism for event accumulation in time, to maintain the time correlation between spikes. In the related works, to achieve good performance, the time correlation is maintained with different methods:

\begin{itemize}[leftmargin=*]
    \item Histograms of Averaged Time Surfaces (HATS)~\cite{N-cars_dataset}: it uses local memory to compute the average of Time Surfaces, which represents the recent temporal activity within a local spatial neighborhood. 
    \item Hierarchy Of Time Surfaces (HOTS)~\cite{HOTS}: it uses the computation of Time Surfaces in a hierarchical way between the layers.
    \item Gabor-filter~\cite{gabor_filter}: it considers the spatial correlation between different events and assigns them to the channels based on this information.
\end{itemize}

In HATS~\cite{N-cars_dataset}, all approaches are evaluated by a simple linear Support Vector Machine (SVM) classifier on the N-CARS dataset. The results of this simple classifier method are compared with our \textit{CarSNN} in Table~\ref{tab:results_comparison}. The Gabor-filter method adopts a two-layer SNN before the SVM. As discussed in Section~\ref{subsection:method_training}, since the upper bound of $T_l$ is $10\ ms$ for the real-time constraint, the comparison is made taking into account this limitation.

\begin{table}[h]
    \centering
    \caption{Comparison of results for $T_l=10\ ms$.}
    \begin{tabular}{c|c}
        Classifier (Accumulation approach)  &  $acc._{test}$\\ 
        \hline
        Linear SVM (HOTS)& $\approx 0.54$ \\
        Linear SVM (Gabor-SNN) & $\approx 0.66$ \\
        Linear SVM (HATS) & $\approx 0.81$ \\
        \textbf{CarSNN} ($128 \times 128$ attention window) & $ 0.86$\\
        \textbf{CarSNN} ($100 \times 100$ attention window) & $ 0.86$\\
        \textit{CarSNN} ($50 \times 50$ attention window) & $ 0.79$\\
    \end{tabular}
    \label{tab:results_comparison}
\end{table}

As highlighted in Table~\ref{tab:results_comparison}, our \textit{CarSNN} achieves better accuracy with a limited $T_l$ than the Linear SVMs implemented after the use of different and more complicated accumulation approaches.

\section{Conclusion}\label{section:conclusion}

In this work, we present \textit{CarSNN}, a novel SNN model for the \textit{``cars vs. background''} classification of event-based streams implemented on neuromorphic hardware. With a three-step process, the network model, training parameters, and input parameters are defined. An attention window mechanism is proposed to accumulate the events focusing the attention on the region in which the majority of the events occur. Two versions of our \textit{CarSNN} with different attention windows achieve 86\% accuracy (drops to 83\% when implemented onto the Loihi Chip), with only $0.72\ ms$ latency, in the worst case, which is 5\% higher than the previous state-of-the-art approaches with an upper bound of $10\ ms$ latency. Moreover, considering also the power consumption of only $315\ mW$ for its implementation on the Loihi Neuromorphic Chip, our \textit{CarSNN} establishes as a prominent method for embedded real-time classification, and opens new avenues toward resource-constraint efficient AD applications on neuromorphic hardware.

\section*{Acknowledgments}

This work has been partially supported by the Doctoral College Resilient Embedded Systems, which is run jointly by the TU Wien's Faculty of Informatics and the UAS Technikum Wien. This work was also jointly supported by the NYUAD Center for Interacting Urban Networks
(CITIES), funded by Tamkeen under the NYUAD Research Institute Award CG001 and by the Swiss Re Institute under the Quantum Cities™ initiative, and Center for CyberSecurity (CCS), funded by Tamkeen under the NYUAD Research Institute Award G1104.

\begin{refsize}
\bibliographystyle{ieeetr}
\bibliography{main.bib}
\end{refsize}

\end{document}